\pdfoutput=1
\documentclass[10pt]{article}
\usepackage[a4paper,margin=.75in]{geometry}
\usepackage[T1]{fontenc}\usepackage[utf8]{inputenc}
\usepackage{lmodern,microtype,amsmath,amssymb,booktabs,graphicx,array,longtable,pgfplots,needspace}
\pgfplotsset{compat=1.18}
\usepackage[hidelinks]{hyperref}
\hypersetup{pdftitle={When More Planning Hurts: Learned Planners for World Models},pdfauthor={Josef Chen},pdfsubject={Empirical evaluation of learned action updates, executed task outcomes and compute cost},pdfkeywords={World Models, learned planners, JEPA, robotics, physical AI, PushT, adaptive computation}}
\usepackage{helvet,xcolor,caption,titlesec,placeins}
\definecolor{paperink}{HTML}{26333B}
\titleformat{\section}{\sffamily\large\bfseries\color{paperink}}{\thesection}{.7em}{}
\titleformat{\subsection}{\sffamily\normalsize\bfseries\color{paperink}}{\thesubsection}{.65em}{}
\newcommand{\pe}{planning plus encoding}
\begin{document}
\begin{center}
{\LARGE\sffamily\bfseries When More Planning Hurts\\[3pt]
\Large Learned Planners for World Models\par}
\vspace{6pt}\normalsize Josef Chen\quad\small Independent Researcher\; \href{https://orcid.org/0009-0004-3651-7391}{ORCID}\\[4pt]
\small 25 October 2026
\end{center}
\begin{abstract}On PushT, our fixed action updater (1,505 parameters) solves 8 of 16 scenes reserved from our configuration selection. The comparison planners perform as follows: Selected Adam (7 tasks solved, total planning and encoding time 218.440s), adapted iCEM (2 tasks solved, total planning and encoding time 49.309s), released LePlanner (16 tasks solved, total planning and encoding time 5.652s). The total planning and encoding time for our action updater is 59.079s. Note that our action updater does not satisfy the task-cost criterion we fixed before evaluation.
We also conduct another matched experiment where we only replay the first ten commands in each context and then allow the next solve 4, 8 or 12 updates (subsequent solves all get 8 updates). Among all 16 contexts (8 scene clusters), the 3 budgets get 9, 8 and 7 successes, respectively. Interestingly, the extra updates did not rescue any failing task in this experiment. All the changes in success/failure happen in one scene cluster, where we observe that the extra updates decrease the costs of the selected latent world-model trajectories for our updater, while the tasks that used to be successes become failures.
The code, weights, evaluators, execution traces and cost accounting accompany the paper.
\end{abstract}
\section{Introduction}
Returning to the simulated PushT visual manipulation task, we perform a matched intervention study: from the same observed history, we revisit the same planning point, but vary the amount of effort expended. We study a small, supervised action updater and ask: (RQ1) can we use a single fixed updater across related predictors and horizons? (RQ2) does the updater improve task success and cost relative to conventional and previously released learned controllers? and (RQ3) given a particular planning decision, do additional updates yield a better result? RQ1 is a test of compatibility, whereas RQ2 and RQ3 are tests that require executing the chosen actions.

\begin{figure}[!ht]
\centering\includegraphics[width=\linewidth]{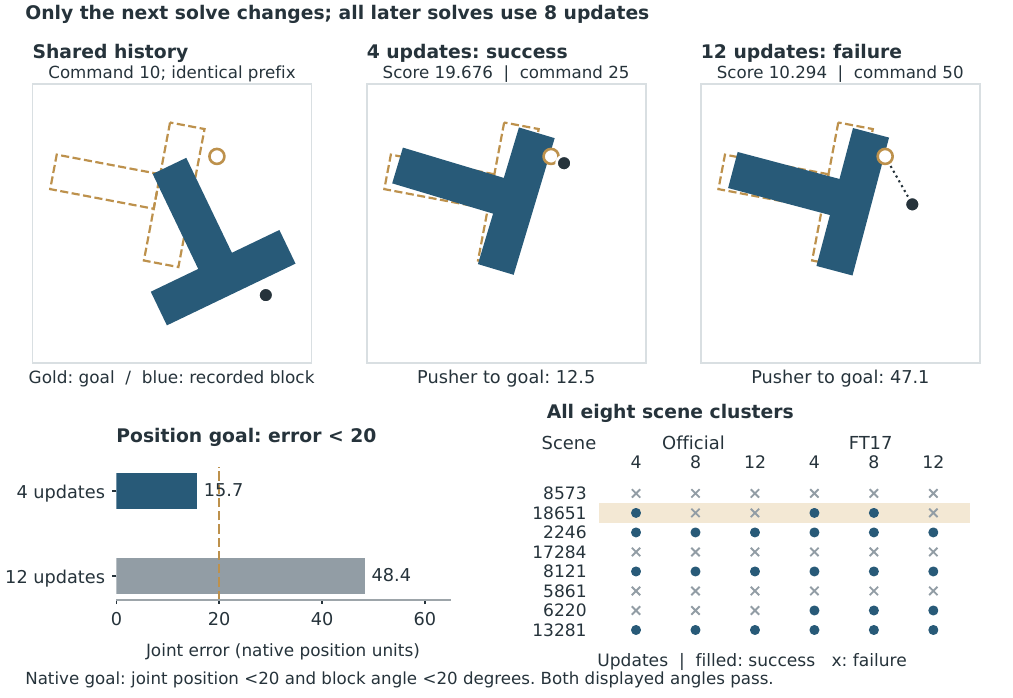}
\caption{\textbf{Similar block poses can hide a failed goal.} In scene 18651, the official predictor gets a ten-command history and follows the plans with four and twelve updates. For the plan with four updates, the task succeeds at command 25. For the plan with twelve updates, the selected score is lower and the task fails at command 50. This is the only solve that differs between these two settings; in all other solves, eight updates were used. Success requires matching not only the block position and orientation, but also the pusher position. The recorded poses at the end of the episode are used to check the pusher's distance from the goal and the joint position error. The two displayed angle errors are both within tolerance. In the figure, gold marks the goals in the dataset; blue and black show the recorded block and pusher poses, respectively; dotted lines indicate the pusher position error. All results in the matched experiment for every scene/model/effort combination are shown in the bottom panel. The scene in which there is a difference is highlighted.}
\label{fig:recorded-task}
\end{figure}

Recall that plans with better predicted scores can still fail at execution time. Our matched intervention tells us which planning decision changes the outcome, but it does not tell us why lower scores precede failure. In RQ3, our additional updates cannot recover tasks, and even make success rate worse in one scene. In the reserved comparison for RQ2, our updater does not satisfy the predefined task and cost criterion and does not perform better than the previously released LePlanner. The poses in these failures are shown in Fig.~\ref{fig:recorded-task} with the goal conditions they failed.

\begin{table}[!ht]\centering\small
\caption{Findings and their scope. We report each experiment separately. In particular, additional controller runs within a scene are not additional scene units.}\label{tab:claim-evidence}
\begin{tabular}{p{.20\linewidth}p{.38\linewidth}p{.33\linewidth}}\toprule
Question & Observed evidence & Interpretation limit\\\midrule
Task value & Reserved sixteen-scene confirmation: updater $8/16$, LePlanner $16/16$; updater fails the predefined criterion & Since the controllers vary on more than just the controller architecture, we can't draw conclusions about the controller architecture independently of population effects\\
Value of effort & Matched $K\in\{4,8,12\}$: $9/8/7$ successes; all changes in one of eight clusters & We do not test learned allocation policies. Also, since only one scene shows a change in outcome, conclusions about this are limited\\
Reuse & One frozen head runs across three related predictors and three horizons & While all predictors share the same encoder and controller interface, it's unknown whether the test scenes appeared in their pretraining data, so it's difficult to make strong conclusions about transfer\\\bottomrule
\end{tabular}
\end{table}

We present our results with failures, cold costs and original sampling unit for each experiment. We reserved the continuation scenes before execution as a training/calibration pilot, on which we do not fit any allocator. We do not use confirmation scenes for configuration selection, but we do not know if these were seen in public-model pretraining. Our initial development experiments, conducted on the original 24 scenes, are reported in Appendix~\ref{app:development}, which also tabulates the full set of numerical results. Results for our reserved comparison are discussed in Section~\ref{sec:confirmation}.

\FloatBarrier
\section{The updater fails the task--cost criterion}\label{sec:confirmation}
\begin{figure}[ht]\centering
\includegraphics[width=\linewidth]{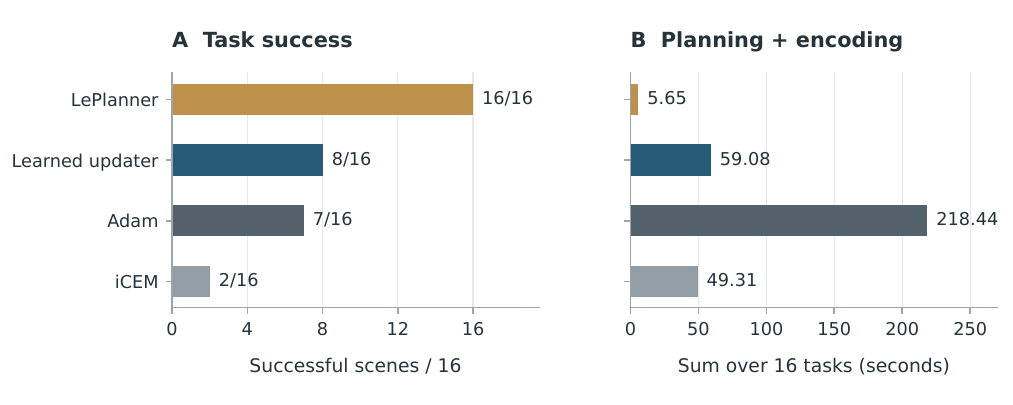}
\caption{Sixteen-scene confirmation. Shown are (left) number of successful tasks (out of sixteen total scenes) and (right) total planning and encoding time (in seconds; note that the sum includes unsuccessful tasks and the cold cost), using linear axes with different units, for four controllers that use different computations. In terms of the practical criterion, the updater fails against more traditional control approaches, and note that here LePlanner is being assessed as a complete controller.}
\label{fig:confirmation}\end{figure}

We perform all 64 assigned trials on these scenes, which are the same sixteen reserved scenes (Table~\ref{tab:confirmation}). The released version of LePlanner solves 16/16 and achieves the best observed task-cost performance among the tested controllers (Fig.~\ref{fig:confirmation}). Our eight-update model (\texttt{learned8}) solves 8/16, selected Adam solves 7/16, and selected iCEM solves 2/16. LePlanner solves the same eight tasks as our updater and also solves the other eight. LePlanner’s summed \pe{} time is 5.652 seconds, versus 59.079 seconds for our model.

Our updater does not satisfy the prespecified criterion. This criterion is not satisfied by counting only the tasks it rescues. Compared with Adam, our updater rescues four tasks and misses three other solved tasks (net $+1$, less than $+3$), though it uses 0.270 times as much summed planning plus encoding time. Compared with iCEM it rescues seven tasks and misses one other solved task (net $+6$, at least +3) but takes 9.770 seconds more in summed planning plus encoding time (ratio 1.198, not at most one).

\begin{table}[ht]\centering\small
\caption{The confirmation study consists of 64 trials in total, using four fixed controllers on sixteen reserved
scenes. Plan trajectories and backward calls are also included, but they measure different aspects of work, so they are not directly comparable across controllers. The time cost of planning and encoding (P+E) is summed together. For LePlanner, we sum its inclusive timer only once. All costs, including failures and cold starts, are included. Task time includes P+E.}
\begin{tabular}{lrrrrrr}\toprule
Controller & Success & Commands & P+E (s) & Task (s) & Plan traj. & Backwards\\\midrule
Adapted iCEM500 & 2/16 & 725 & 49.309 & 54.317 & 73,177 & 0 \\
Adam32, lr 0.1 & 7/16 & 621 & 218.440 & 220.676 & 4,318 & 4,064 \\
Fixed learned8 & 8/16 & 603 & 59.079 & 61.493 & 1,230 & 984 \\
Released LePlanner & 16/16 & 243 & 5.652 & 6.715 & 216 & 0 \\
\bottomrule\end{tabular}
\label{tab:confirmation}
\end{table}

\FloatBarrier
\Needspace{12\baselineskip}
\section{Controller and evaluation protocol}
\subsection{Objective, update and carried plan}
The terminal objective is
\begin{equation}
 J_m(A\mid h_t,g)=\|\mathcal R_{P_m}(h_t,A)_H-E(g)\|_2^2,
 \label{eq:objective}
\end{equation}
where $g$ is the goal image, $h_t$ denotes the observation/action history available to the agent prior to making a decision, and $a_j\in\mathbb R^{10}$ denotes an action block containing 5 two-dimensional (2D) raw commands represented in normalized coordinates. Here, $E$ is the frozen encoder, $P_m$ is a predictor with the same interface, and $\mathcal R_{P_m}$ is the autoregressive rollout from $P_m$. We denote $A=(a_1,\ldots,a_H)$ as a sequence of $H$ action blocks.
Note that the two approaches also differ in terms of implementation and in terms of the variables being optimized over. In the case of the original five-knot approach, five 10-coordinate knots are decoded to the desired horizon, while in the case of the direct-action approach, the optimization is performed over $10H$ active coordinates (30/50/70 active coordinates in the case of horizons H3/H5/H7, respectively). Hence, a comparison of the two approaches does not allow us to isolate the effect of the architecture.

The shared head $u_\theta:\mathbb R^{12}\to[-1,1]$ has layer widths $12\to 32\to 32\to 1$, SiLU hidden activations, a tanh output activation, and 1,505 parameters. Equation~\ref{eq:update} shows the update for the $j$-th scalar coordinate of the complete plan (not the action-block index $j$ from above).
\begin{align}
 x_j^k&=\phi_j(A^k,\nabla_AJ_m(A^k),J_m(A^k),\Delta A^{k-1},r_k),\nonumber\\
 A_j^{k+1}&=\Pi_{[\ell_j,u_j]}
 [A_j^k+\rho(u_j-\ell_j)u_\theta(x_j^k)].
 \label{eq:update}
\end{align}
The head does not receive the scene/model identifier, hidden physical state, contact information, future actions or a success label. Within a solve, we cache the current/history/goal embeddings. The 12 channels are: bounded action, RMS-normalized gradient, gradient sign, bounded cost, relative sample cost, normalized horizon position, log horizon, log block dimension, remaining query fraction $r_k$, previous within-solve update, cost rank and log gradient RMS. Single-candidate relative cost and rank are inactive.

$A^*$ is always one of the members of $\mathcal S_K$, where $\mathcal S_K$ includes the visited candidates, the initial candidate (which has been scored), and the final candidates that meet the query budget. A final update that is not scored is thus never chosen. Among the members of $\mathcal S_K$, our optimizers with a fixed query budget return the candidate plan $A^*$ with the lowest objective $J_m$. After executing one five-command block, the next solve will start from the shifted plan
\begin{equation}
 \Phi(A^*)=(a_2^*,\ldots,a_H^*,0).\label{eq:shift}
\end{equation}
In addition, the next solve takes as input the updated real images and the action history (executed actions so far). The zero tail of the shifted plan is in normalized coordinates. Note that the previous-update features refer only to the current solve, but the plan is carried over from one solve to the next.

\subsection{Fitting and exposure}
Our pooled fit uses 144 (correlated) snapshots and thus 5,760 coordinate examples, and is trained for 100 epochs (2,300 Adam optimizer updates), at seed 13. For reproducibility, we use the pinned versions of the public inputs~\cite{lewmassets,pushtdata}. Note that different coordinates do not correspond to independent tasks.
To imitate objective improvement, we learn $u_\theta$ to minimize $\mathcal L(\theta)=N^{-1}\sum_i(u_\theta(x_i)-y_i)^2$ over $(x_i,y_i)$ pairs, where $y_i\in[-1,1]$ is a normalized teacher-improvement target (obtained from optimization with the frozen model). Our approach does not use RL with task rewards.

\paragraph{Exposure and interpretation.} Pilot scenes are assigned before seeing results, and related predictors share the action interface and encoder.
\begin{description}\setlength{\itemsep}{1pt}
\item[Updater TRAIN.] Trains the updater on official and FT17 H3/H5 teachers (FT11/H7 not used for updater fitting, though not excluded from world-model pretraining).
\item[Original comparisons.] Three predictors and horizons H3/H5/H7 across 24 explored scenes; exploratory paired reuse.
\item[Frozen gate EVAL.] Four additional scenes with 16 ID and 20 challenge contexts; fixed fits and frozen-policy screen.
\item[Conventional TRAIN/CAL.] Four exposed calibration scenes and four held-out validation scenes; six preassigned settings, success-first family selection, no retuning after confirmation.
\item[Common confirmation.] Sixteen scenes, official H5 predictor, four frozen controllers. Scenes are reserved for confirmation and not used for our selection; we cannot rule out that they were in public pretraining.
\item[Continuation TRAIN/CAL.] Eight fresh scenes, official and FT17 H5 predictors, and efforts fixed at 4, 8, and 12 updates. Prospective pilot; no fitted allocator, no confirmation.
\end{description}

\subsection{Controls, execution and statistical unit}
An episode terminates either upon successful completion of a task or after a maximum number of 50 executed commands. To determine whether or not a task has been successfully completed, we query the environment both at the beginning of a planning step and after the execution of a command in both our controller and the released controller.
World/CEM solver probe: during planning we pause the physics simulation and don't use hidden physics states as features for the controller, they are only used for reset/evaluation and for replay. The probe stores its initial evaluation, which happens at the first command executed after solving. The history contains up to three images as well as the actions taken.

Continuation experiment setup. We keep the same 513-query ceiling, use $\rho=0.1$, and vary only the number of optimization updates at the first continuation decision (immediately after the shared prefix) and fix subsequent continuation decisions to 8 updates. The release also includes seeds, details of the normalizer, full per-setting bounds, and other operating points.

In the original development comparisons, different controllers perform different computation under the common 513-query ceiling. With bounded CEM (64 candidates, eight elites), one CEM iteration plus initialization uses 65 forward trajectories per solve. Projected Adam uses 12 steps and learning rate 0.3. With 12 updates, the shared head uses 14 forwards and 12 backwards. The shared head optimizes Eq.~\ref{eq:objective}; task success is measured separately. For PushT, success requires the joint pusher/block-position error to be less than 20 task coordinate units and the block-angle error to be less than $\pi/9$, where the angle comparison accounts for the symmetry of the block shape. Figure~\ref{fig:recorded-task} illustrates why block alignment alone does not determine this outcome.

Relationship to the published LeWM PushT controller. The published LeWM PushT controller~\cite{lewm}, App.D uses 300 CEM candidates, up to 30 CEM iterations with 30 elites, H5/frame skip 5, and executes all 25 planned commands before replanning. It is worth noting that the context size used during training is not necessarily the same as the context (history) size used during deployment. For example, although our evaluation harness allows a history size of up to 3 images, the released public planning YAML used a history size of 1 image. Similarly, for PushT, we use a replanning cadence of five commands and a smaller search budget. Therefore, changing the population size alone cannot reproduce the published controller; the CEM numbers in our table correspond to our controller, and are not an exact reproduction of the reported LeWM PushT numbers.

\subsection{Reserved-scene comparison and task--cost criterion}
All methods use the same official LeWM predictor (H5), the same five-command cadence, the same allowed actions, and the same reset/goal interface; we compare full controllers under these shared conventions and a budget of at most 50 commands. For LePlanner, we use the released setup: its training/support/arrival-hold behavior, three refinement steps, returning the final decoded plan, the released normalizer, the 6.8M-parameter controller, query initialization, and zero-action padding for early latents. We evaluate on all 16 originally reserved scenes and do not further tune controller hyperparameters on these scenes; we do not know whether these scenes were included in the authors' pretraining of the released public model. We evaluate the four controllers in a fixed order (the selected iCEM500 controller, the selected Adam32/lr 0.1 controller, the unchanged eight-update model, and released LePlanner), for a total of 64 controller-scene trials. All three of our methods use the same terminal objective and normalizer (iCEM also retains the distribution/elite memory).

This success/time criterion was set before looking at the experimental outcomes: learned8 must outperform each selected conventional comparator in net successes by at least three, spend no more total planning and encoding time than that comparator, and complete all assigned trials. Net successes subtract learned8-only task failures from learned8-only task rescues; LePlanner, evaluated separately on the same scenes, is not included in this comparison. For decision latencies, we report nearest-rank descriptive quantiles over individual decisions rather than scene-level aggregates; the first cold decision of each trial is included, and so are decisions in ultimately failed tasks.

Core percentile intervals (descriptive 95\% conditional on fixed fits, related predictors and the exposed set; ignoring multiplicity and not accounting for refit/model-family uncertainty) were calculated using 20,000 seeded scene resamples, each of which preserved all predictor/horizon/method cells of each of 24 scenes. Continuation intervals similarly preserved all arms and predictors within each of eight clusters. The number of branches did not increase the effective number of independent samples. Note that these intervals are not evidence of equivalence/noninferiority even if they include zero.

\Needspace{10\baselineskip}
\section{Matched continuation experiment}
\subsection{One encountered solve, fixed subsequent policy}
For RQ3, we replay eight new scenes with FT17 and the official predictor at H5, resulting in 16 scene-predictor pairs. This allows us to study the effect that the amount of compute spent on a decision has on eventual task success, holding constant the history of decisions made previously. For each pair, we use the eight-update controller to take ten commands. As before, we first ensure that the replay behavior matches that of uninterrupted execution, and each arm separately reconstructs the environment reset, command/render order, image/action history, normalized incoming plan and random state. At this point, none of the episodes has completed the task. We vary the solve corresponding to command 10, which uses $K\in\{4,8,12\}$, and all later solves use eight updates. All 48 continuations run without fitting.

Here $Y_c(K)$ is eventual success under this intervention and the fixed policy used after it, and hence $D_c(K',K)=Y_c(K')-Y_c(K)$ is the executed difference in success when we intervene on effort for that solve (as opposed to the remaining effort in the episode, or the predictor/updater). Note that membership and ordering of arms are fixed ex ante (before outcomes are seen), and that only information available at runtime is used (e.g.\ not hidden physical state, contact, success or future labels).

\begin{table}[ht]\centering\small
\caption{All three efforts;16 contexts in eight scene clusters per row.
Counts/times include one acquired prefix plus each suffix, excluding replay.
Every later solve uses 8.}\label{tab:effort}
\begin{tabular}{lrrrr}\toprule
First updates&Successes&Forwards&Backward/NN calls&P+E sum(s)\\\midrule
4&9/16&1146&904&55.983\\8&8/16&1290&1032&62.309\\
12&7/16&1404&1136&68.310\\\bottomrule\end{tabular}
\end{table}

Extra effort rescues no task. K8 loses one official success relative to K4;
K12 loses one FT17 success relative to K8. Both losses occur in scene 18651. None of the three update budgets we test outperform the baseline of always using four updates. For instance, even adaptively selecting the effort after seeing each result only achieves a 9/16 success rate, precisely the success rate obtained by always using four updates. Any choice that keeps every four-update outcome also leaves no saving in forward passes, backward passes or neural network calls. Across the 16 contexts considered (grouped into 8 scene clusters), none of the update budgets we test offer any success or call-count advantage over always using four updates. We leave further controlled experiments to quantify how often additional effort is counterproductive for task execution.

\subsection{Lower selected scores precede failure}
\begin{figure}[ht]\centering\input{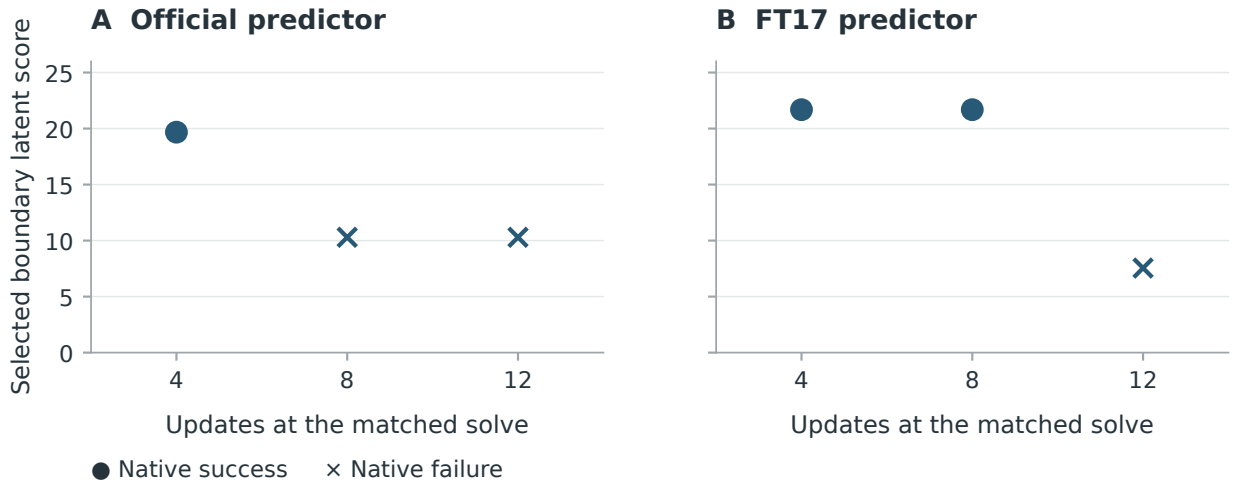}
\caption{Scene 18651: selected score and eventual outcome. Filled circles indicate success and crosses failure. Scores can only be compared within the context of the same model. All binary changes occur in this one cluster out of eight.}\label{fig:score}
\end{figure}

Official/scene 18651 K4 selects cost 19.6764 and succeeds at command 25; K8/K12
select 10.2942 and fail at 50. With FT17 in the same scene, the task succeeds at command 22 using K4/K8 with cost 21.6850. Using K12 selects 7.5485, and the task fails at command 50. Figure~\ref{fig:score} shows both affected contexts. In both cases, more updates give the lower score and lose the task. This may be due to inaccuracies in the dynamics predictor, the metric geometry, contact/horizon effects, unsupported actions, or later decisions in the fixed policy.

\subsection{Whole-policy controls and accounting}
CEM solves 7/16 and Adam solves 5/16 tasks (planning and encoding time per task averaged over all tasks is 0.669s and 6.573s respectively, with sums of 10.710/105.175s) when controlling whole policies with different controller histories (each using their own controller prefix) on the same set of tasks/scenes/models. CEM is much more computationally efficient but it is not conclusive which algorithm is better due to similar success rates. The later conventional planning/control comparison uses a different set of four scenes (for development validation).

The total number of forwards and backward/NN calls included in the 16 acquired prefixes is 320 and 256 respectively. Replay performs 480 reconstruction commands in 1.982 seconds. Note that these are part of the experiment overhead, and not to be confused with deployment planning. The total execution time (including startup, execution, capturing and writing the artifacts) is 197.675/205.551 seconds for the experiment/control, respectively. Note that we should not add the nested execution time again. Table~\ref{tab:effort} shows the eager prefix-plus-suffix cost instrumented in the paused simulation. For completeness, the full counts, failures and cold costs are released with the results.

\section{Related work}
Our work also relates to latent planning methods. PlaNet plans with rewards predicted through stochastic latent dynamics~\cite{planet}. DINO-WM~\cite{dinowm} and V-JEPA2~\cite{vjepa2} use pretrained visual models: DINO-WM plans toward goal embeddings using distance in the pretrained feature space, while action-conditioned V-JEPA2 performs image-goal robot planning by minimizing L1 energy in latent space. Our LeWM backbone is trained from offline pixel/action trajectories with prediction and SIGReg losses and plans by minimizing the squared latent distance between the terminal prediction and the goal embedding~\cite{lewm}, \S3. The stable-worldmodel platform standardizes model/solver/evaluation interfaces and examines broader environmental shifts~\cite{swm}, \S\S3--4. In this paper we focus on the goal-distance objective and use it to study related latent predictors, different updaters, and planning horizons.

Learning the update rule is established. Coordinate-shared optimizers reuse
parameters across optimizee dimensions~\cite{l2o}, \S2.1. Sacks and Boots train
joint MPC distribution updates from warm distributions and structured-sample
costs using learner-visited teacher corrections~\cite{sacks}, \S III. DMPO
learns residual optimizer and shift models through executed reward with a
cost/distribution actor interface~\cite{dmpo}, \S III. RP-1 also learns a plan-refinement operator conditioned on the current plan, terminal value and its action gradient~\cite{rp1}, \S4, Eqs.3--8. Its goal-conditioned critic is learned with offline temporal-difference targets, and its planner is trained through imagined rollouts. Our updater imitates teacher improvements of a terminal latent-distance objective. Our experiments test this recipe; RP-1's reported performance is not reproduced here. LePlanner uses goal residuals and rollouts from a frozen model to improve its plan tokens, with objectives for action-support and arrival-and-hold that are distinct from terminal-cost imitation~\cite{leplanner}, \S\S3.2--3.6. Other planners or controls have mechanisms that are specific to them. For example, Adam keeps track of moments of the gradient~\cite{adam}, MPPI can shift its control sequence between solves~\cite{mppi}, \S4, and iCEM can have temporal noise, shifted and retained elites, and execute the best-seen action sequence~\cite{icem}, \S3. Our tested updater has a feedforward coordinate head, teacher-improvement objective, and fixed plan shift. We test these iCEM mechanisms in a bounded adaptation; MPPI remains untested. Note that just naming an algorithm CEM does not imply either mechanism.

There are also related studies which investigate adapting the compute allocation in other ways. MPC meta-policies learn both the prediction horizon and when to recompute~\cite{metampc}, \S III. The method ``Stop! Planner Time'' learns hyperparameters and when to continue vs.\ execute based on contextual information, optimizing for the utility of the terminal policy~\cite{stop}, Algorithm 1. When to Plan makes use of contextual information including task progress, uncertainty, and history to choose between a reactive controller and planning-horizon options~\cite{whentoplan}, \S\S4--5. AdaReP adapts thresholds on when to refresh a plan based on mismatch and sensitivity~\cite{adarep}, \S4.2. DeepJEPA chooses how many internal transition prediction steps to make based on the marginal improvement in latent prediction, while keeping the outer CEM loop fixed~\cite{deepjepa}, \S3. SufficientPlan calibrates a fixed budget for each model and task from paired executed outcomes, and reuses static context within solves~\cite{sufficientplan}, \S4. Our intervention changes the number of outer-loop updates at one encountered planning call, and we do not fit an allocator in this test. There is precedent for budget non-monotonicity in SufficientPlan, which reports success at a reduced budget even when full planning fails.

The term objective mismatch has at least two distinct meanings: the mismatch between the dynamics model's training objective (likelihood) and the controller's objective (reward maximization) discussed in~\cite{mismatch}, \S\S3--4, and the mismatch between improving some performance metric during optimization (taking descent steps w.r.t.\ a fixed objective) and the resulting plan succeeding at the task eventually. Different methods address different flavors of the mismatch problem: MOPO uses a penalty on outcomes predicted by the dynamics model with high uncertainty in order to avoid exploiting inaccuracies in the offline regime~\cite{mopo}, \S4, value-guided JEPA regularizes the representation by shaping its geometry using an objective based on the reaching value~\cite{valuejepa}, \S3, and LePlanner uses a combination of a support measure penalty and arrival/hold penalties. When a controller's score decreases and the task is subsequently failed it is unclear whether the root cause is a) the dynamics model makes inaccurate predictions, b) the controller explores actions outside the support of the data, or c) optimizing the chosen objective does not lead to a successful controller. We report paired controller outcomes and quantify uncertainty where possible given the tested fits, but acknowledge that care must be taken when making inferences about coverage of the intervals and generalization performance, as more empirical evaluation is required to validate such claims. We believe that first agreeing on what the unit of analysis/sampling is and ensuring that the baselines used are competent will improve the reliability of empirical results~\cite{rliable,empdesign}.

System-One decision models pick among explicit options. For example, TypeSafe describes its model Jev as returning typed probabilities in parallel, trained using Reinforcement Learning for Calibrated Decisions (RLCD)~\cite{typesafejev}. In a recent Jev study, Jev chooses among hand-designed robot skills based on PD control and inverse kinematics, and inverse-kinematics solutions are screened using copied physics~\cite{jevphysics}, \S6 and App.E. That work separated evaluating supplied consequences from predicting consequences. We don't evaluate either Jev or RLCD; our updater imitates improvements in terminal objectives. Our results for continuation motivate asking whether it's worth doing additional deliberation steps to get better results when executing, before learning an allocation policy.

\section{Discussion and limitations}
The tested updater achieves task success on eight of the sixteen reserved scenes and does not satisfy the task-cost criterion against conventional controls; released LePlanner solves all tasks on these scenes. Compatibility with different horizons and related predictors is not evidence of task success. Comparisons also change controller capacity, fitting data, objectives, normalization and initialization, so they do not isolate the effect of architecture. The confirmation scenes have been used and will not be used for any further tuning.

In our case study we look at two related predictors and eight scene clusters. We found instances where making additional updates decreases the selected score; in one scene it also decreases the number of successes. There are a variety of reasons this might happen (inaccurate predictor, geometry of latent metric, action support, contact effects, planning horizon, downstream decisions, etc.), and a single example isn't enough evidence to support a general mechanism. We would need at least multiple matched examples or a controlled change to the objective/predictor model. Four updates performs as well as the outcome oracle, and we did not find an update budget that both uses fewer calls and maintains the successes of four updates.

Several aspects remain unexplored, including public pretraining exposure, robustness across more fitting seeds, and transfer to an independent encoder/dynamics family. In addition, all experiments target a single task with related predictors and fixed control cadence. Finally, the cost-only control setting is limited to corrective learning with four rounds, unlike the DAgger regime with 20 rounds considered by Sacks and Boots. Thus, we do not rule out the possibility that their full approach is effective. MPPI was not tested.

While the appendices contain more details on cost boundaries, historical comparisons, and full numerical results, we encourage readers to not over-interpret our results: (i) we do not claim any practical utility of the learned allocator, (ii) the updater should not be considered a new optimizer family, nor (iii) should the results be taken to generalize to other model families. For reproducibility we make failures, weights, and configs (and when available: plan/neural traces) available. Only a subset of experiments/replays were verified for execution. We leave it to future work to rerun the full experiment matrix, e.g.\ including real robots.

\section*{Research assistance}
StealthGPT was used to make suggestions for improving the writing of this manuscript. Some of the suggested changes were incorporated after careful checking to ensure consistency with the original methods and results. Generative AI tools assisted with coding, literature searches, analysis and manuscript drafting during this project. Pangram was used for text classification.
\clearpage\begingroup\small\bibliographystyle{plain}\bibliography{references}\endgroup
\clearpage\appendix
\section{Compatible reuse and development comparisons}\label{app:development}
For RQ1 we use the same (frozen) head for each predictor/horizon cell with a shared interface, and for RQ2 we study its task success and cost. In the direct-action cohort, we find that 140/216 tasks can be solved using either shared or padded features (Adam solves 132/216 tasks and CEM solves 119/216 tasks). The success rate difference between shared and Adam is $+3.70$pp $[-6.48,+13.89]$ and the mean difference in total planning plus encoding time is $-0.458$ seconds $[-1.053,+0.052]$, with a mean of 5.832 seconds for shared and 6.289 seconds for Adam. Predictor rollouts are not padded, but note that with and without padding the head's outcomes agree on all 216 binary success/failure cases.

\begin{table}[ht]\centering\small
\caption{The operating points for the direct-action methods correspond to 216 tasks per method from 24 scenes, three horizons and three predictors. We report numbers including all failures and cold work. Note that episode time includes the time spent in planning/encoding, so do not add the two.}
\begin{tabular}{lrrr}\toprule
Method & Native goals & Plan + encoder (s) & Episode (s)\\\midrule
CEM & 119/216 & 1.029 & 2.912 \\
Adam & 132/216 & 6.289 & 8.227 \\
Shared SFT & 140/216 & 5.832 & 7.732 \\
Same head, padded & 140/216 & 5.831 & 7.673 \\
\bottomrule\end{tabular}
\label{tab:direct}
\end{table}
\begin{figure}[ht]\centering\input{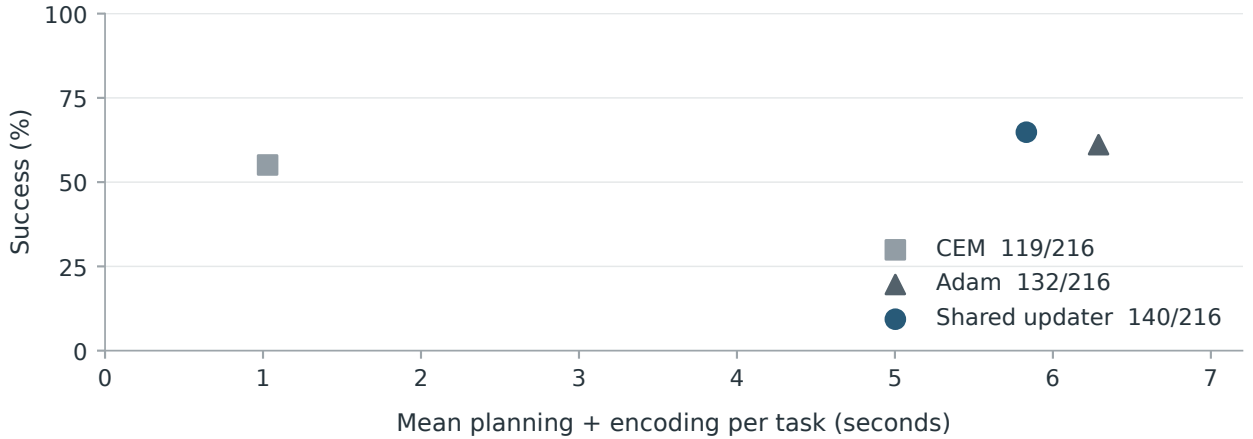}
\caption{Success and cost of direct-action at the configured settings, with instrumentation included in the means. Fig.~\ref{fig:paired} gives the paired intervals.}\label{fig:operating}
\end{figure}

The compiled shared updater takes $+1.681$ seconds $[-0.765,+5.000]$ more than compiled Adam on average: mean planning plus encoding time in the compiled cohort is 3.321 seconds for the shared updater, and 1.640 seconds for Adam. The paired intervals leave unresolved whether Adam or the shared updater has the joint advantage on cost and task performance. In the compiled cohort, the shared updater solves 128/216 tasks, Adam solves 122/216, and CEM solves 106/216. The paired difference shared-minus-Adam is $+2.78$pp $[-7.41,+12.50]$. In the original five-knot cohort, the shared updater solves 129/216 tasks, and Adam solves 122/216. There are 38 tasks solved by only the shared updater, and 31 tasks solved by only Adam. The paired difference shared-minus-Adam is $+3.24$pp $[-6.94,+12.96]$. Mean planning plus encoding time in the original five-knot cohort is 5.362 seconds for the shared updater, and 5.324 seconds for Adam. The total number of tasks solved is the same for compiled and eager Adam, although there is one task whose outcome flips in each direction. CEM keeps the same outcome on all 216 tasks.

\begin{figure}[ht]\centering\input{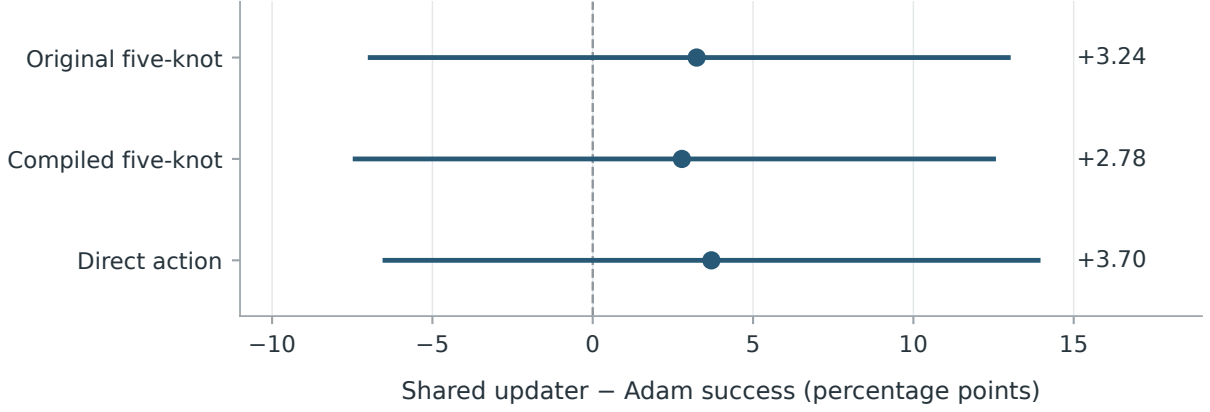}
\caption{The figure shows shared-minus-Adam success differences with descriptive 95\% scene
intervals. The scenes are the same 24 for each cohort. Comparisons across cohorts should be interpreted with caution since the parameterization and implementation vary between cohorts. As a result, observed differences are not necessarily attributable to a single factor.}
\label{fig:paired}\end{figure}

The four fixed specialists, each trained on a different cell, solve 110, 135, 141 and 127 of the 216 tasks, while the pooled updater solves 140. The best specialist thus leads by one goal ($+0.46$pp $[-8.33,+8.80]$), but note that teacher exposure and the amount of fitting work are not matched. Using seeds 13, 23 and 37 for fitting (conditional on using the same teacher corpus) leads to 140, 131 and 141 tasks being solved. When FT17 adaptation is cached, using the warm head leads to 35/48 of the target goals being solved, just as when using the fresh or pooled heads, and 33/48 of the official retention set being solved (32 when using the original unchanged head and 34 when using the fresh or pooled heads). Removing gradient magnitude and scale during fitting and inference (while still using the gradient sign) leads to 110/216 ($-13.89$pp $[-24.07,-3.70]$) and forgetting the retained plan leads to 63/216 ($-35.65$pp $[-45.37,-25.93]$). These experiments give some indication as to what features of the gradients the updater may be leveraging as well as the importance of the retained plan, but they do not establish new warm-start methods.

Compared to fixed8, the earlier episode-start threshold ties on primary goals (9/16 solved), with faster planning (46.518 s vs 50.158 s, 7.2569\% saving). However, it did not pass the screen because it did not solve two more primary goals without increased planning cost, or achieve at least a 25\% planning-time reduction without giving up any solved primary goals (the screen also required no worse service). On the 20 challenge contexts, both solve 8, with slightly higher cost for the threshold. For reference, the random policy solved 9/16 primary goals (31.847 s planning) and the ridge policy 4/16 (5/20 on challenges). The appendix includes the full results tables, including all failures.

\Needspace{12\baselineskip}
\subsection{Released-controller and World/CEM solver development probes}
Table~\ref{tab:baseline-probes} separates out the four-scene probe from its common two-scene subset. The only controller to solve all four scenes is the released LePlanner; learned8 and Adam each solve two, and bounded one-step CEM only one. On these four scenes, LePlanner's summed planning/encoding time is 6.072 s, versus 16.713 s for learned8, and it issues 60 vs.\ 154 commands. Thus, LePlanner solves a superset of tasks relative to the shared updater. These planners/controllers differ in capacity, training, objective, normalization, and plan state.

\begin{table}[ht]\centering\small
\caption{Two additional probes were also run for development purposes, on a four-scene and on a two-scene cohort. In all cases, the numbers are cumulative over the whole set of completed tasks. The number in the P+E column includes planning and encoding (and transfer when available). Plan trajectories is the number of plans (i.e.\ action sequences) considered and not the number of module calls (i.e.\ the computational cost can be different even when the numbers are the same). Module and transition counters are still present in this release. Note that P+E is included in the task service number. Hence, no extra column for it should be added. Cold-start, method-order and pipeline differences remain.}
\begin{tabular}{p{.34\linewidth}rrrrr}\toprule
Method/protocol & Success & Commands & Plan traj. & P+E(s) & Task(s)\\\midrule
\multicolumn{6}{l}{Four development scenes: 10646,2690,6916,2424}\\
Released LePlanner & 4/4 & 60 & 56 & 6.072 & 6.931 \\
Owned learned8 & 2/4 & 154 & 310 & 16.713 & 17.267 \\
Projected Adam & 2/4 & 148 & 420 & 22.073 & 22.576 \\
Bounded one-step CEM & 1/4 & 167 & 2210 & 2.341 & 2.875 \\
\midrule
\multicolumn{6}{l}{Common two-scene subset: 10646,2690}\\
Released LePlanner & 2/2 & 35 & 32 & 5.436 & 6.199 \\
Owned learned8 & 0/2 & 100 & 200 & 11.538 & 11.898 \\
Projected Adam & 1/2 & 75 & 210 & 11.367 & 11.620 \\
Bounded one-step CEM & 0/2 & 100 & 1300 & 1.489 & 1.797 \\
Native CEM: history 1/cadence 25 & 2/2 & 48 & 18000 & 5.025 & 5.734 \\
Native CEM: history 3/cadence 5 & 0/2 & 100 & 180000 & 52.636 & 53.564 \\
\bottomrule\end{tabular}
\label{tab:baseline-probes}
\end{table}

We do not reproduce the benchmark results reported in the original papers, since under our protocol both the success rate and the cost are affected. On the two scenes we have in common, LePlanner (and World/CEM with history 1/cadence 25) solves 2/2, while World/CEM with history 3/cadence 5 solves 0/2. In total, World/CEM spends 5.025/52.636 s in encoding/planning under the two different protocols, searching 18,000/180,000 candidate trajectories. Note that we cannot infer the cause of this performance difference, since history length and cadence differ at the same time. The controls returned by the World/CEM solver are unbounded (one reaches absolute value 1.026411 at time step 40 on scene 10646), and the final reported elite-average score is not the score of the mean action, so it is not directly comparable to our best-scored-plan selection rule.

The interrupted World/CEM solver acquisition is reported separately in Appendix~\ref{app:baseline-binding}.
The first two scenes in Table~\ref{tab:baseline-probes} were also used in our smoke run. These are not separate, independent samples; they are development probes that occur before calibration and reserved-scene confirmation.
The first LePlanner task takes 5.802 seconds to solve out of the 6.931 seconds to solve all four tasks. We do not subtract this as a ``warmup'' because the loading and publication of the model as well as controller-related operations are measured in separate scopes.

\Needspace{11\baselineskip}
\subsection{Matched conventional calibration and separate validation}
\begin{figure}[ht]\centering
\includegraphics[width=\linewidth]{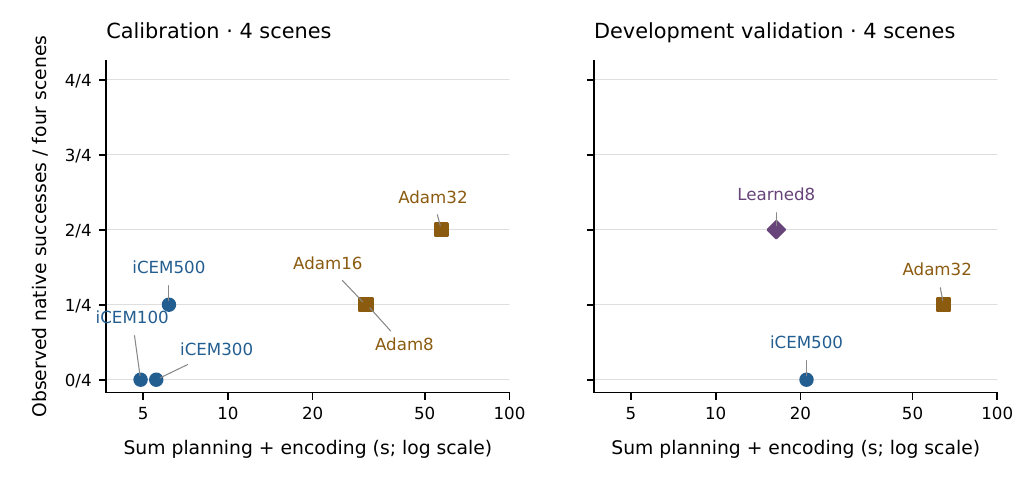}
\caption{Bounded conventional calibration (left) and separate development
validation (right), four scenes per panel. The horizontal axis is the
completed-task sum of planning plus encoding in seconds, on a logarithmic
scale; the vertical axis is the task success count out of four scenes.
All nine settings and all failure cases are presented, with the points representing instrumented executions. The two scene sets remain separate.}
\label{fig:conventional}\end{figure}

On bounded calibration, the method with the highest success count within each family, namely Adam32/lr 0.1 and iCEM500, was selected (Table~\ref{tab:conventional}), with no timing tiebreak. We then ran learned8 and the selected methods on a hold-out set of four validation scenes (different from the 16 scenes used for confirmation). Learned8 solved 2/4 scenes, the selected Adam method solved 1/4 scenes, and the selected iCEM method solved 0/4 scenes. Specifically, learned8 is the only method that solved scene 1245, learned8 and Adam solved scene 7842, and all three methods failed on scenes 6138 and 17184. The summed planning plus encoding times for learned8, the selected Adam method, and the selected iCEM method were 16.417, 64.410 and 21.011 seconds, respectively. Figure~\ref{fig:conventional} shows performance on calibration and validation scenes separately. Learned8 performed the best on the validation scenes, and the confirmation on the 16 scenes then tests it against the pre-defined criterion.

\begin{table}[ht]\centering\small
\caption{Note that Plan trajectories is the number of action plans evaluated during planning, but does not necessarily correspond to the amount of computation, as different algorithms make a different number of calls to the backward method and/or batch in different ways. Task service includes P+E. P+E is total planning plus encoding time across all decisions the agent makes while executing the task (including cold starts and instrumented work). Each row corresponds to four scene units, with calibration and validation on different scenes, in a matched experiment with conventional controls.}
\begingroup\setlength{\tabcolsep}{8pt}
\begin{tabular}{lrrrrr}\toprule
Setting & Success & P+E (s) & Task (s) & Plan traj. & Backwards\\\midrule
\multicolumn{6}{l}{Calibration: four exposed scenes}\\
Adam8, lr 0.1 & 1/4 & 31.187 & 33.402 & 350 & 280 \\
Adam16, lr 0.3 & 1/4 & 30.782 & 31.398 & 630 & 560 \\
Adam32, lr 0.1 & 2/4 & 57.278 & 57.902 & 1,122 & 1,056 \\
iCEM100 & 0/4 & 4.897 & 5.621 & 3,988 & 0 \\
iCEM300 & 0/4 & 5.566 & 6.331 & 11,908 & 0 \\
iCEM500 & 1/4 & 6.176 & 6.742 & 16,148 & 0 \\
\midrule
\multicolumn{6}{l}{Validation: four separate development scenes}\\
iCEM500 & 0/4 & 21.011 & 23.149 & 20,188 & 0 \\
Adam32, lr 0.1 & 1/4 & 64.410 & 65.064 & 1,156 & 1,088 \\
Fixed learned8 & 2/4 & 16.417 & 17.047 & 300 & 240 \\
\bottomrule\end{tabular}

\endgroup\label{tab:conventional}
\end{table}

\Needspace{10\baselineskip}
\section{Development controllers and selection}\label{app:development-methods}
\subsection{Additional controller and World/CEM solver probes}
We further evaluate on four additional official/H5 development scenes. We compare our learned8 against the released version of LePlanner~\cite{leplanner}, projected Adam and bounded one-step CEM baselines. The released LePlanner consists of 6,802,445 parameters in the controller, repeats the early context to three frames, performs three refinements, and initializes learned plan queries each solve. Our controllers instead shift the previous action plan and grow real history from one to three images. The released controller uses its own saved action normalizer, whereas ours use the original normalizer. All of these controllers use the exact same official world-model bytes. Note that this compares complete controllers, which means that in addition to differing in terms of model capacity and training/data, they also differ in terms of auxiliary objectives (arrival/hold training and support objective~\cite{leplanner} in the released controller), normalization, and persistent plan state. It is unknown to us whether the four additional development scenes have been used in training the released controller.

In addition, we have run the first two scenes also with a World/CEM 300-candidate/30-iteration CEM protocol (with 30 elites and variance 1) with two combinations of history/cadence (1/25 and 3/5), where history and cadence vary together. In all cases actions are unbounded and we take the final mean of the elites returned by the solver (note that the reported elite-average cost is not a cost we have scored for that mean on its own). We have used caching of the static encodings and we have followed our (adapted) local protocol for the tasks, normalizer and randomization. Note that we have not reproduced the published benchmark quality/latency. We count the work for the completed tasks and the interrupted acquisition separately.

\subsection{Matched conventional calibration and development validation}
In this later bounded comparison, we keep the terminal objective, the official predictor at H5, the original action normalizer, the number of real images ($\le 3$), the reset/goal construction, the cadence (five commands) and the maximum number of commands (50) fixed across all settings. Raw action bounds are $[-1,1]$. All settings run on four exposed calibration scenes. We compare Adam with different numbers of steps (8/16/32) and learning rates (0.1/0.3/0.1). For iCEM, we try three combinations of initial population size (40/100/150) and number of iterations (3/4/5). For all iCEM settings, we use 10 elites, temporal colored noise with $\beta=2.5$, population decay 1.25, momentum 0.1, and retain a 0.3 fraction of elites~\cite{icem}. We report the nominal number of trajectories (100/300/500), but the settings actually score 97/295/502 trajectories at the first solve and 100/298/505 in later solves. Retained scores from the same context are reused; shifted elites are freshly scored. The returned plan has a measured score. iCEM's initial sampled actions are at the center of the action box, and it fits and shifts a distribution with a repeated tail. This is different from our approach that initializes from the carried-over plan.

The frozen rule selects one configuration from each family by success count, breaking ties using total planning plus encoding time, number of trajectories, and then configuration ID. We apply the chosen configurations (and learned8 as-is) to four different preassigned TRAIN/CAL scenes (12 trials total) without further validation-time retuning. This serves as a development validation (distinct from the 16 confirmation scenes used later). We do not report a reproduction of the iCEM paper's published results; our iCEM is an adaptation to our setting, and compute is not equalized across arms (while the 513-trajectory cap is the same, the amount of compute that is actually used need not be). We retain all 24 calibration trials. Note that model capacity and fitting history as well as initialization and random seed are also not controlled for across arms.

\section{Experimental records and accounting}
For each run, we track the executed commands, the objective and cost at each decision step, the task outcome, and the initial state and goal. Every number in a table can be traced back to specific runs (and specific versions of source code and model weights). Note that different cohorts of runs may have been generated with different limits on sampling and measurement, so when looking at the same cases multiple times, don't treat each readout as additional independent evidence---each cohort of runs should be interpreted using the limits that applied at the time.

The fresh whole-process startup cost is 32.244 seconds for the assay and 81.514 for controls. (This is not additive with the process total of 197.675/205.551 seconds but nested inside it; the complete readout keeps eager instrumentation, clones/diagnostic input hashes and renderings.) The total for the prefix simulator is 16.607 seconds, also nested.
Success is determined according to the success criterion of the task at hand, not according to the progress towards the block-center (which is reported separately).

\section{Missing comparisons and tests that could overturn the findings}
\begin{description}\setlength{\itemsep}{2pt}
\item[Released LePlanner~\cite{leplanner}.] The released LePlanner includes normalization and was trained using consequence-aware refinement, with arrival/hold behavior and action support. We ran experiments on four development scenes, as well as all sixteen common confirmation scenes (using the saved normalization); however, the individual contributions of the various elements in its controller are still not clear.
\item[iCEM/MPPI~\cite{icem,mppi}.] For iCEM/MPPI, temporal noise, retained sampling information, and warm-started distribution updates can also influence interpretation. So far, we calibrated three adapted iCEM budgets on four scenes and tested the selected iCEM and Adam on four development and sixteen common confirmation scenes. However, we did not include MPPI in our experiments.
\item[Published LeWM planner~\cite{lewm}.] Note that we have not been able to replicate the quality of the benchmark in LeWM, and have only run two scenes using two adapted LeWM planner protocols. The interpretation of our results might be affected by using the official LeWM planner (with 300 candidates, up to 30 iterations and the full 25-command execution).
\item[Sacks/Boots and DMPO~\cite{sacks,dmpo}.] Sacks/Boots and DMPO can in principle change the interpretation via a sufficient teacher correction, or via a reward-trained update/learned shift. At the moment we don't have evidence for either of those effects: our short imitation control does not reproduce them, and we haven't run matched, sufficient training.
\item[Fixed-budget certification~\cite{sufficientplan}.] Formal certification on the population level is still an open issue. In the fixed-budget setting, task-outcome calibration may already be able to explain the observed benefit without knowing the details of the allocation. Current evidence: calibration, separate development validation, and frozen confirmation were performed; the margin of the updater was not met.
\item[Cheap contextual policies~\cite{adarep,whentoplan,stop}.] It's also worth noting that our current evidence is that the earlier threshold idea didn't pan out and that higher effort didn't actually get us more successes than just using fixed~4. So cheap contextual policies could potentially change the interpretation, given that allocation outcomes can be explained by history/progress/mismatch signals and gating overheads.
\item[Independent model/task shift.] However, these experiments do not investigate whether the benefits of re-use generalize beyond using similar weights and a single task. This could be interesting to study in a follow-up work, e.g.\ by training a completely independent model, or performing a task shift, as these could alter the interpretation of the findings. The encoder's public-data exposure also remains.
\end{description}

\section{Baseline records and acquisition costs}
\label{app:baseline-binding}

Controller/worker wall time is reported as aggregate totals over all tasks. Per-task controller/worker wall time by algorithm is intentionally not summarized because the fixed method order can make wall time partly reflect instrumentation/cold-cache effects. In the four-method LePlanner comparison controller wall time is 64.062 s and worker wall time is 61.604 s. World/CEM's observed command was 1.026411.
To facilitate reproducibility, we provide provenance information including released source versions, hashes of the checkpoints used to obtain the data and pointers to the exact files and fields we used to extract the data from. For our baselines, we use the sixteen completed task records from four scenes as well as the four task records obtained from the World/CEM solvers on a two scene subset (re-using the same scene identities as in the completed task records). One of these tasks exited prematurely with exit code 143 (reason unknown), and we reran this task using the same reset, solver and source code, and let it run within the remainder of the 900 second time limit. We verified that the first 9 solver evaluation records and the first 45 commands obtained match those from the interrupted task exactly. We count such tasks once in the denominator of four trials in the campaign.
Worker/controller wall time for rerunning this task was 32.173/42.702 seconds (this work isn't free). We do not infer the total amount of time taken by the original four-trial campaign (based on the sum of the three completed tasks).
For this one task we know that we obtained at least 86,100/430,500 additional candidate trajectories and transitions beyond the completed tasks (198,000/990,000), so in total we obtained at least 284,100/1,420,500 candidate trajectories/transitions (the actual numbers could be larger since we don't know the full extent of the additional data from the incomplete task). Costs for the public pretraining stage are not reconstructed.
World/CEM reporting here includes an elite mean and an elite-average score, but not a rescored mean-action objective. The payload omits the first observation after reset.
Although the number of candidate trajectories and training/acquisition exposure differ from method to method, results are bounded and auditable via predicates, commands, and goal/state information, along with records of the work performed by each module. Conventional baselines don't serialize actual past-action tensors; reconstructed action blocks are validated independently of LePlanner's saved-input checks.
The oracle, counterexample, differences from prior work and source-format checks do not establish deployed benefit. To rule out a learned effort interpretation it would suffice to show that (i) the observed operating points (in the sense of a trade-off between task and cost) can be explained by a fixed or cheap policy that performs well; (ii) the observed matched effects disappear over repeated trials; or (iii) in reserved scenes the observed benefit cannot be predicted based on the information available before making the effort-related decision. For claims of broad transfer, evidence on at least one additional independent task or system family is required. Similarly, claims of future noninferiority require that the margin be specified in advance.

\section{Conventional-control evidence and what the timers include}
Table~\ref{tab:conventional} reports results on all 36 records from tasks with conventional controls, and the family-selection rule is fixed. Calibration scenes are 10646, 2690, 6916 and 2424; validation scenes are 1245, 6138, 7842 and 17184. For the earlier LePlanner probe, we used the same calibration scenes, but it did not introduce any additional independent calibration scenes, nor did we measure performance on validation.

Each of the two supervised calibration and validation campaigns respectively took 324.529 and 255.116 seconds wall-clock time to run to completion. For the two campaigns respectively, worker wall-clock time was 311.538 and 166.382 seconds, and worker startup time was 167.543 and 59.834 seconds. This time is included in the above scopes and should not be added to task wall-clock time or P+E wall-clock time. Both campaigns were run on a local GTX1660Ti GPU (float32; Torch 2.8/CUDA 12.6) with 2 CPU threads, and no extensions were requested. As these numbers include cold start / instrumentation costs (since we keep all trials that ran to completion), do not use these as an estimate of time-to-completion on an actual robot. (In addition, the simulator is paused while planning.)

\section{Confirmation records}
Table~\ref{tab:confirmation} was obtained using the 64 confirmation task records, which also yield the paired rescues/losses, cost ratios and practical criterion discussed above. We include all necessary inputs in the source package and outline the procedure to reproduce the results. The confirmation worker/supervisor required 478.218/528.888 seconds (this includes startup time of 130.398 seconds). Do not include these times in computing the P+E time and task time. The time to make the first iCEM decision (22.090 seconds) is included in iCEM's cost. Lastly, the paused simulation is instrumented, so these numbers do not represent production latencies.

\section{Complete historical numerical record}
All the experiments listed below have been done separately, any differences in execution times between experiments should not be interpreted as speedups. All results and error estimates are quoted as originally computed.
\section{Original comparison and compiled conventional controls}
For methods in the fixed-five-knot matrix we report results aggregated over all 216 cases (1,296 episodes in total). Failures remain in the denominator. In terms of the adequacy results, the learned control (A115) for cost-only performs poorly compared to the network-free best65 selector (56/216 cases solved as opposed to 114/216, paired difference: $-26.85$ pp, descriptive scene interval $[-41.20,-12.50]$). This is in contrast to its numerical fit results.

\begin{table}[ht]\centering\small
\caption{Original fixed-five-knot matrix, showing the mean time (in seconds) for each episode. Note that the definition of an episode varies by source, and also that the instrumentation is not the same in all cases.}
\begin{tabular}{lrrr}\toprule
Method & Native goals & Plan + encoder (s) & Episode (s)\\\midrule
Cost-only A115 & 56/216 & 1.056 & 4.247 \\
Halton--Boltzmann & 116/216 & 0.532 & 2.053 \\
CEM & 106/216 & 0.550 & 1.928 \\
Adam & 122/216 & 5.324 & 6.795 \\
Shared SFT & 129/216 & 5.362 & 6.711 \\
best65 & 114/216 & 0.690 & 3.828 \\
\bottomrule\end{tabular}
\end{table}

In comparison to Adam, there are 38 tasks that SFT solved when Adam didn't and 31 tasks that Adam solved that SFT didn't (difference in number of tasks solved: net seven; $+3.24$pp $[-6.94,+12.96]$). In terms of total planning and encoding time, SFT took on average 5.362 seconds whereas Adam took 5.324 seconds (paired difference: $+0.038$ seconds $[-0.293,+0.394]$). Thus, we can't conclude that one had a robust advantage over the other in terms of cost or performance (or that they were equal). The conventional controls, which were much cheaper, have lower cost as expected.

\begin{figure}[ht]\centering
\includegraphics[width=\linewidth]{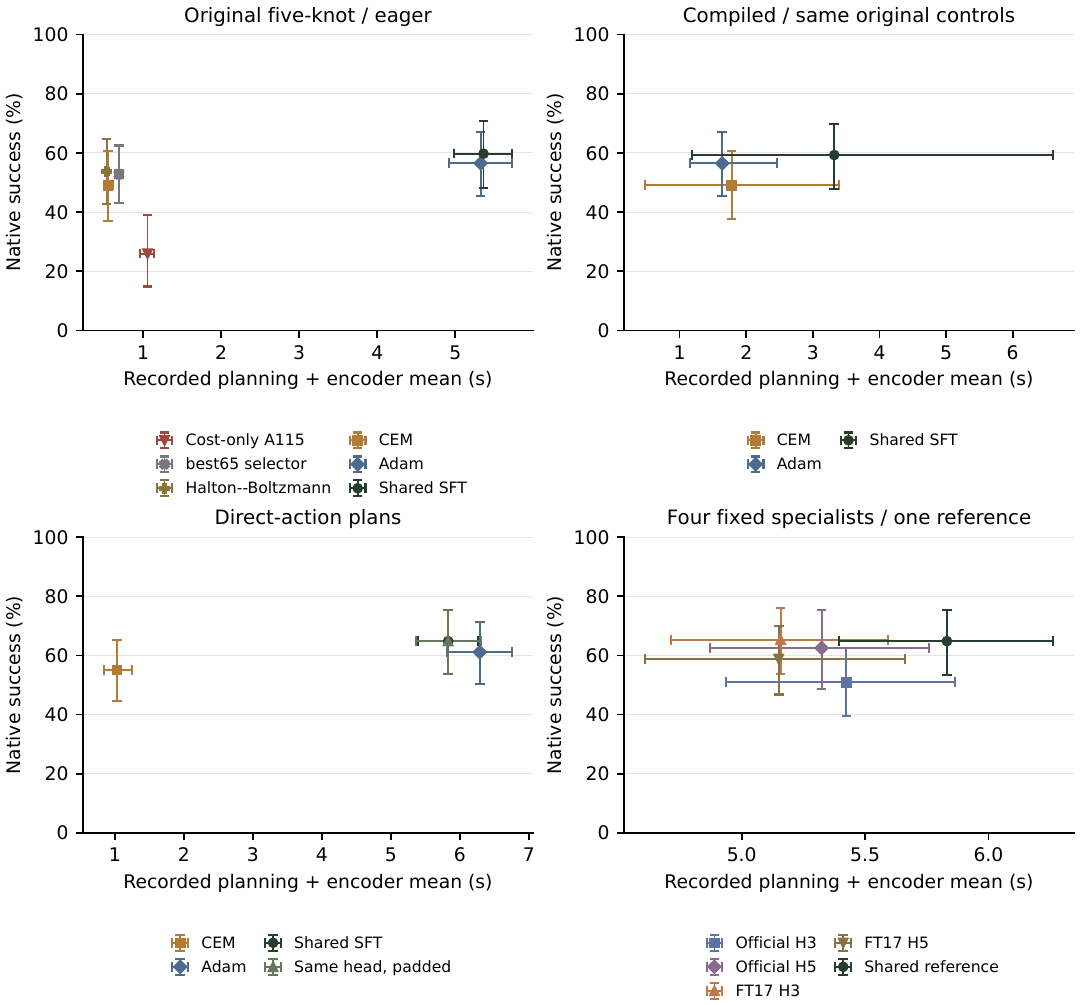}
\caption{For each cohort, the panel summarizes the range of task-cost. The details of parameterization, recording and timing vary between cohorts. Note that these intervals correspond to resampling of source scenes and do not represent resampling of decisions or model fits.}
\end{figure}

This doesn't mean that using compiled mode with Adam always makes it faster, nor does the fact that the number of solved tasks is the same mean that they both use the same commands.
Compiled mode also didn't change Adam's number of solved tasks: it solved 122/216 tasks in both modes (though one paired task switched from solved to unsolved and another one switched the other way). CEM's per-pair solved statuses also agree for all 216 tasks.
Note that we do not subtract the cost of compilation in ``cold'' runs, and we did not retune any of the methods on TEST. Compilation time in cold runs, as well as failed preparation attempts, are thus included in these numbers.
On a compiled matrix of 648 cases (the same original task/config/seed identities as before, all with reduce-overhead enabled), SFT solved 128/216 tasks, Adam solved 122/216 tasks, and CEM solved 106/216 tasks. The difference Shared - Adam is $+2.78$pp $[-7.41,+12.50]$. Shared SFT took on average (planning + encoding time) 3.321 s, whereas Adam took 1.640 s (difference $+1.681$, $[-0.765,+5.000]$). For Adam, using compiled instead of eager mode changes its (planning + encoding time) by $-3.684$ s $[-4.325,-2.822]$.

\begin{table}[ht]\centering\small
\caption{Compiled controls, run to completion, with 216 trials recorded for each method.}
\begin{tabular}{lrrr}\toprule
Method & Native goals & Plan + encoder (s) & Episode (s)\\\midrule
CEM & 106/216 & 1.780 & 3.218 \\
Adam & 122/216 & 1.640 & 2.984 \\
Shared SFT & 128/216 & 3.321 & 4.658 \\
\bottomrule\end{tabular}
\end{table}

\section{Direct-action parameterization and fixed specialist transfer}
In a separate study with direct actions with a total of 864 episodes, Shared and padded shared succeed on 140/216 episodes, Adam 132 and CEM 119. The paired difference shared-minus-Adam is $+3.70$pp $[-6.48,+13.89]$. The paired difference in total (planning plus encoding) time shared-minus-Adam is $-0.458$ $[-1.053,+0.052]$ seconds (6.289 for Adam, 5.832 for shared). Here, for padded shared, we padded the feature rows (to 70) but not rollouts, and this control has the same number of parameters. All 216 paired binary outcomes agree between shared and padded shared. The success count (140 vs. 129 in the previous study) differs, and the two studies differ in parameterization, implementation and data recording. This study extends the previous one by using direct actions with four different methods on the same 24 source scenes and with three predictors (H3/H5/H7).

\begin{table}[ht]\centering\small
\caption{Direct-action H-dimensional controls. Padding retains the same fitted head.}
\begin{tabular}{lrrr}\toprule
Method & Native goals & Plan + encoder (s) & Episode (s)\\\midrule
CEM & 119/216 & 1.029 & 2.912 \\
Adam & 132/216 & 6.289 & 8.227 \\
Shared SFT & 140/216 & 5.832 & 7.732 \\
Same head, padded & 140/216 & 5.831 & 7.673 \\
\bottomrule\end{tabular}
\end{table}

\subsection{Four fixed specialists, one historical reference}
We do not perform head selection based on TEST performance, and there is only one run in the shared setting with 216 cases, which serves as the reference for all comparisons. In order to compare the transfer performance of the 4 heads (trained on the four official/FT17 H3/H5 cells) we report results on all 216 original direct-action cases. For each head 24 of these cases are aligned with its training cell and 192 are transfer cases, for a total of 96 aligned and 768 transfer episodes.

Specialists FT17-H3/H5 solved 141/127 of the 216 problems, while Official-H3/H5 specialists solved 110/135, and the shared reference solved 140. The one-goal difference in favor of FT17-H3 equates to $+0.46$pp $[-8.33,+8.80]$.
The aligned specialist/reference goals are 17/19, 15/15, 19/19 and 16/16 of the 24 total, while their transfer goals are 93/121, 120/125, 122/121 and 111/124 of the 192 total.
Note that these results depend on the particular (fixed) training cell that we use, i.e., having the favorable head does not correspond to a routing policy learned during training.

\begin{figure}[ht]\centering
\includegraphics[width=\linewidth]{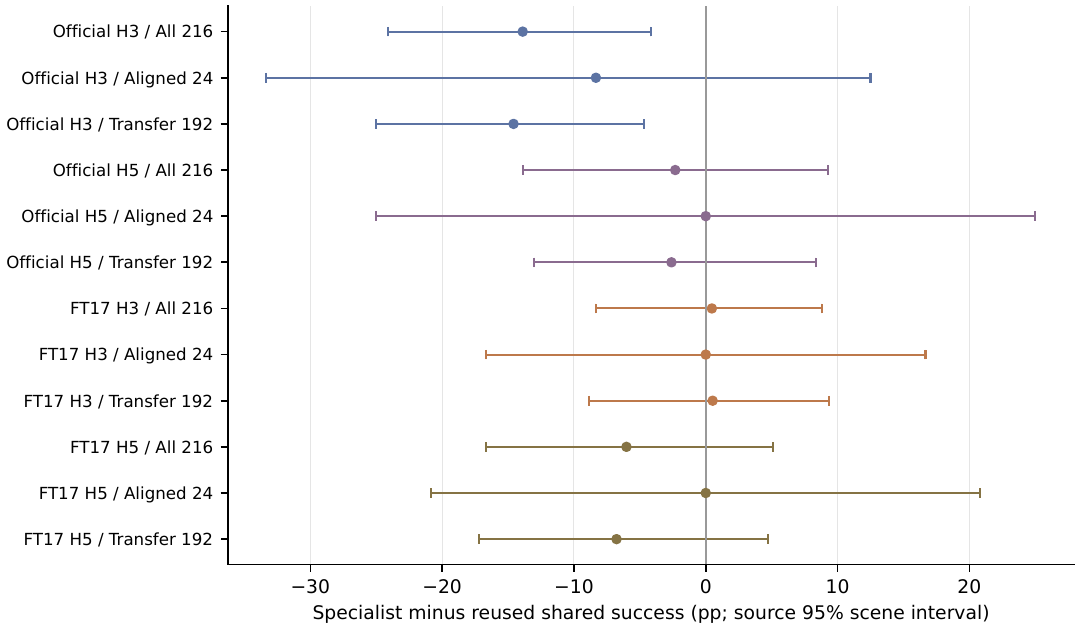}
\caption{For all 24 aligned and 192 offdiagonal partitions, intervals are computed conditional on the fitted heads and related predictors. Note that since all comparisons are made to the same historical shared reference, these comparisons are not independent. Positive values favor the specialist. Each fixed specialist is compared to a common reference (in a predeclared order). These intervals are computed for all 216 cases.}
\end{figure}

\begin{table}[ht]\centering\small
\caption{All cells (task/model/H) reported according to the source records. There are 216 evaluations for each fixed specialist and the reused pooled reference model.}
\begin{tabular}{lrrr}\toprule
Method & Native goals & Plan + encoder (s) & Episode (s)\\\midrule
Official H3 & 110/216 & 5.424 & 6.402 \\
Official H5 & 135/216 & 5.324 & 6.377 \\
FT17 H3 & 141/216 & 5.158 & 6.068 \\
FT17 H5 & 127/216 & 5.150 & 6.087 \\
Shared pooled & 140/216 & 5.832 & 7.732 \\
\bottomrule\end{tabular}
\end{table}

All heads have 1,505 parameters and are trained for 100 epochs, but specialist H3/H5 only use 1,080/1,800 coordinates and are updated 500/800 times, unlike the 5,760/2,300 for the pooled version. As such, the amount of data, teacher exposure and amount of fitting work are not comparable. The smaller computation time for the specialists that was reported thus does not demonstrate a total cost saving under comparable conditions nor does it show that specialization is generally more effective than using a shared updater.

\section{FT17 adaptation, fitting cost and official retention}
We compare four roles, fixed in advance of running the study: (i) pooled unchanged, (ii) official specialist unchanged, (iii) official specialist warmed on cached FT17 TRAIN, and (iv) a fresh FT17 specialist. We use 24 scenes and both horizons (H3/H5), with 384 episodes, of which 48 target and 48 retention per role (192 target and 192 retention total). In total we tracked seven unique weight artifacts and eight role/horizon combinations, and report target improvement alongside previously supported-context harm.

\begin{table}[ht]\centering\small
\caption{Complete fixed-role adaptation, with four roles; P+E indicates the recorded mean for planning and encoding. Each role/panel includes 48 task episodes.}
\begin{tabular}{lrrrr}\toprule
Fixed role & Target goals/48 & Retention goals/48 & Target P+E (s) & Retention P+E (s)\\\midrule
Pooled unchanged & 35 & 34 & 3.773 & 3.752 \\
Official unchanged & 34 & 32 & 3.901 & 3.806 \\
Official to FT17 warm & 35 & 33 & 3.854 & 3.867 \\
FT17 fresh & 35 & 34 & 3.718 & 3.673 \\
\bottomrule\end{tabular}
\end{table}

Planning-plus-encoding time averaged 3.854 s in the warm target-goal setting (fresh: 3.718 s, pooled: 3.773 s, official: 3.901 s). Task success for target-goal averaged 35/48 for both warm adaptation and the pooled and fresh baselines, slightly higher than for the unchanged official baseline (34/48; mean difference in success rate for target-goal $+2.08$ percentage points, scene interval from $-12.50$ to $+16.67$pp). For official-goal retention, task success averaged 33/48 for warm adaptation, lower than for the pooled and fresh baselines (34/48) but higher than for the unchanged official baseline (32/48). We emphasize that we do not set any noninferiority margin or adaptation-benefit threshold, so do not make any noninferiority conclusions; the comparisons above should be interpreted as only descriptive indications of direction.

Each warm head receives 100 epochs of FT17 training on top of the official fit (i.e. 500/800 additional FT17 updates, totalling 1,000/1,600, whereas fresh receives 500/800 and pooled receives 2,300 updates). Performance numbers are for separately run tasks. Note that reusing a cached teacher is not the same as having access to the teacher for free. We run Saved-state and full-MSE checks on the initial and final weights, as well as all 36 FT17 TRAIN snapshots for each head.

For direct-action 384, the controller took 1,935.978 seconds, and the workers took 1,846.263 seconds. The two fitter-body timers total 3.894 seconds, within 27.018 seconds of worker time, and 27.636 seconds of total fit-controller time. So far, it looks like in this setting the extra fitting has not paid off for adaptation. (Note: this run is marked as having failed because there was a reaping error 0.1 seconds after EOF, but this was after all evaluations had completed; the numbers include the failed run, but there is no rerun of the failed task to change the status.)

\section{Fit seeds, retrained feature masks and plan reset}
In addition to the original results described above, we release results for two fit seeds, five retrained feature masks and one plan-reset control (seven refits plus the reset control). As before, all original 216 scene/predictor/H cells are included, but we only compare to a single past reference (the seed13 run). The refits all have 1,505 parameters and were run for 100 epochs using 5,760 cached examples and 2,300 updates. In total there are 1,728 episodes, and the release has 81 head/model/H cells and 80 paired contrasts. As before, we include all fixed heads, and the reset control retains the original weights.

\begin{figure}[ht]\centering
\includegraphics[width=\linewidth]{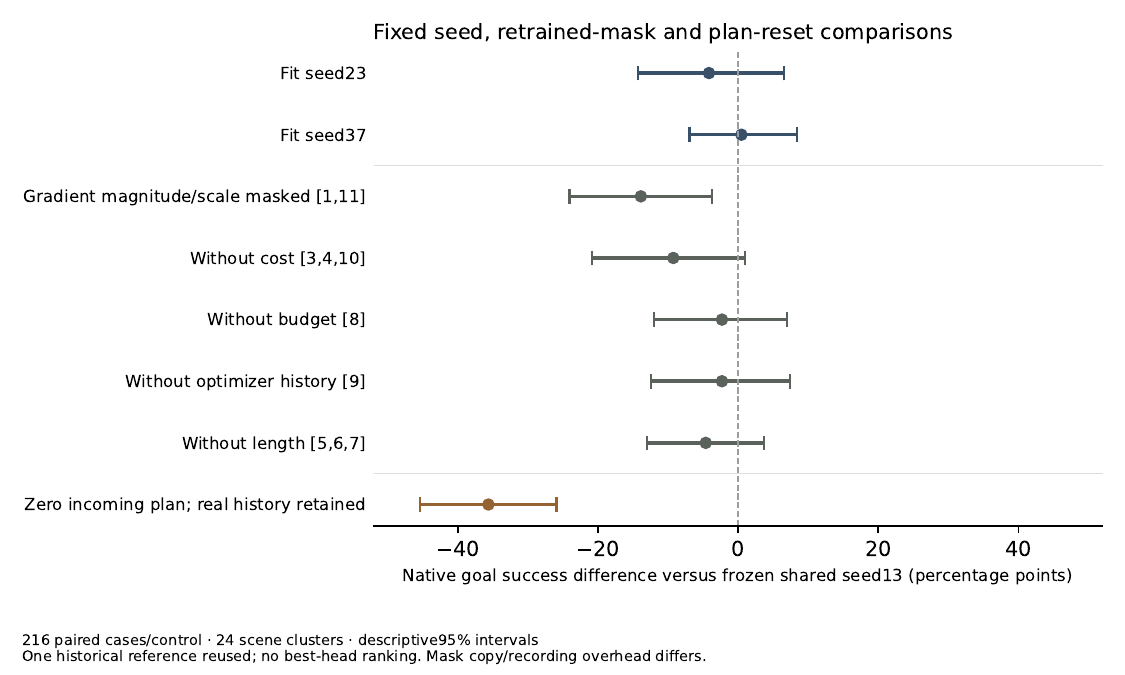}
\caption{All eight fixed modifications relative to the one shared seed13 reference model, with 216 paired tasks each. As before, we use the same 24 scenes and cached teacher lineage, and report unadjusted, descriptive 95\% scene intervals. Notably, removing gradient magnitude and scale still retains the gradient-sign channel.}
\end{figure}

\begin{table}[ht]\centering\small
\caption{Controls are finite and complete, with only a single shared reference being reused.
Mean P+E per episode is left unchanged (preserving the instrumentation and cold costs specific to its source).}
\begin{tabular}{lrrr}\toprule
Fixed head & Native goals & Plan + encoder (s) & Episode (s)\\\midrule
Shared seed13 & 140/216 & 5.832 & 7.732 \\
Shared seed23 & 131/216 & 5.408 & 10.162 \\
Shared seed37 & 141/216 & 5.262 & 9.958 \\
No gradient/scale & 110/216 & 5.589 & 11.120 \\
No cost & 120/216 & 5.437 & 11.049 \\
No budget & 135/216 & 5.351 & 10.441 \\
No update history & 135/216 & 5.365 & 10.606 \\
No length & 130/216 & 5.331 & 10.318 \\
Reset plan each decision & 63/216 & 6.238 & 12.438 \\
\bottomrule\end{tabular}
\end{table}

Fit seeds 13, 23 and 37 give 140, 131 and 141 solved tasks out of 216 (mean 63.58\%, sample SD 2.55pp, range 60.65 to 65.28\%). These results are conditional on one corpus (they are not independent datasets and do not provide a seed interval for a population). Masking the channels related to gradient magnitude and scale $[1,11]$ keeps the gradient sign (channel 2) and leads to 110/216 solved tasks ($-13.89$pp $[-24.07,-3.70]$). Masking the cost channels $[3,4,10]$ results in 120/216 solved tasks ($-9.26$pp $[-20.83,+0.93]$); channels 4 and 10 are already inactive. Masking the budget $[8]$, previous update $[9]$ and coordinate/length channels $[5,6,7]$ results in 135, 135 and 130 solved tasks, respectively, and in these three cases the interval includes 0. Masking is applied during both fitting and inference. Note that an interval that includes 0 does not prove that these channels can be safely discarded.

Resetting the previous solution at each decision step has an even larger negative impact, with performance dropping to 63/216 instead of 140/216 ($-35.65$pp $[-45.37,-25.93]$). Here, the real observed/action history and all 12 within-solve updates are still used. Hence, this setting is not the same as masking the previous-update channel 9. This setting shows that this implementation relies on using the receding-horizon solution as initial solution. As warm-starting is a well-known and established practice, this effect is not regarded as a new algorithm.

Three stage campaigns required 4,525.225, 11,947.411 and 2,794.411 seconds inside a 19,330.396-second controller. The CPU check of the saved results takes 453.082 seconds. Acquisition and fits in cache are accounted separately.

Because the experiments varied in how traces were recorded, and because recording transformed and raw traces adds planning work, we cannot draw conclusions about production latency or the total cost savings under matched conditions. Similarly, we cannot tease out the effect of the fit seed from these changes.

\section{Actual effort allocation and runnable-flow validation}
Five effort policies were chosen before EVAL: (1) fixed8 (always 8 updates); (2) a gradient threshold (4 or 8 updates); (3) random effort independent of task (effort frequencies as in TRAIN); (4) a ridge decision model, which did not pass cross-validation on TRAIN; and (5) fixed8, but with the ridge decision still computed (and ignored). Four separate source scenes supply 16 ID contexts and 20 predictor/horizon challenge contexts, and with 5 effort policies there are 180 complete policy episodes in total. ID contexts: official and FT17 at H3/H5. Predictor/horizon challenge contexts: FT11 at H3/H5/H7, and official/FT17 at H7. Note that effort is selected once per episode using the legal first-prefix data. These are executed policies (not cross-validation recombinations of reset arms from TRAIN).

\begin{table}[ht]\centering\small
\caption{All frozen EVAL policies. Note that the ID/OOD success denominators (16/20) count contexts that repeat over four source scenes. Planning times include the gate.}
\begin{tabular}{lrrrr}\toprule
Actual frozen policy & ID goals/16 & ID planning (s) & OOD goals/20 & OOD planning (s)\\\midrule
Fixed8 & 9 & 50.158 & 8 & 94.635 \\
Threshold4/8 & 9 & 46.518 & 8 & 95.835 \\
TRAIN-marginal random & 9 & 31.847 & 5 & 67.341 \\
Failed ridge gate & 4 & 17.373 & 5 & 26.596 \\
Fixed8 + ignored gate & 9 & 46.519 & 8 & 94.213 \\
\bottomrule\end{tabular}
\end{table}

Here the primary threshold ties fixed8 in number of goals solved (9/16), but does so with 46.518 seconds of planning time as opposed to 50.158 for fixed8, a 7.2569\% saving. Our prespecified screen requires either (a) solving at least two more goals without increasing planning time, or (b) saving at least 25\% planning time without solving fewer goals, and also no worse complete episode service time. The threshold approach has a service time of 71.034 seconds, as opposed to 76.762 for fixed8. It passes this part of the criterion on its own, but still fails the criterion overall because neither of the task/planning time criteria are met. On OOD contexts they tie again, this time by solving 8/20 contexts, with the threshold approach having a planning time of 95.835 seconds as opposed to 94.635 for fixed8. Thus, our held-out test does not confirm the TRAIN allocation result.

All 36 command paths of fixed8 are identical to those of the ignored gate, so any timing difference should not be entirely ascribed to the added scorer computation. Other effects, e.g.\ warm/cold start, execution noise, etc.\ can also play a role. Neither the gate nor the threshold met its prespecified criterion. The failed ridge model solves 4/16 ID and 5/20 OOD goals. Random solves 9/16 ID goals (planning time 31.847s).

\subsection{What the released flow demonstrates}
CEM successfully solved this task (after 41 commands). Adam and shared failed (after 50 commands). The demos include both success and failure examples.
The demo geometry is schematic only; it does not show simulator camera footage or a physical robot.
This is a public task CLI smoke test, including loading and publication (37.869s). It is a quick sanity check, not a re-run of the full 864-case matrix. All three jobs are official/H5/task-slot 0 jobs. Saved-artifact checks cover 141 commands and 120 shared neural calls, and the reproduction package pins exact source and configuration bindings.

\section{Cost boundaries, uncertainty and limits}
Planning, encoding, simulator episode, complete trial, campaign and controller measurements are reported separately, as each measures a different quantity. Note in particular that the simulator episode measurement includes the planning and encoding measurements, and thus timings should not be naively summed together. The timings also do not measure the latency of uninstrumented neural network computation, and are not measurements of response time on a physical robot. For episode execution, timing begins just before creating/resetting the environment and performing rendering, and ends just before closing the environment and saving execution records. In addition to reporting the query ceilings, we report the measured amount of work being done, including recording work, which is measured within the planning timers.

Local experiments: GeForce GTX 1660 Ti; PyTorch 2.8 with CUDA 12.6; float32 control; two CPU threads. Incremental rental cost for local computation is \$0 (we did not track hardware depreciation or electricity costs). All 38 prior closed allocations together have a quoted GPU rental cost of \$18.95 USD, according to our compute ledger. This is our historical acquisition/experimentation cost, and does not represent the total amortized cost of deployment, nor the cost of individual inferences. Note that the exact rental cost may differ as we don't have access to the original invoices and don't know about potential additional fees.

For the three-method run, we used official/H5/slot-0 and a CUDA allocator that reuses memory across the three methods (actual run order: Adam, shared, CEM). Peak allocated GPU memory was 108,029,952 / 108,361,728 / 107,426,304 bytes and peak reserved was 134,217,728 / 136,314,880 / 136,314,880 bytes, for Adam / shared / CEM. Memory usage as measured by the allocator only includes the episode window and as such may contain tensors belonging to models that were already resident as well as tensors cached by the allocator and reused by later methods (e.g. encoding may still need to happen during cold start). It also does not include memory used while loading models, during global device setup, on the CPU, or publishing results. As such, do not interpret these numbers as peak memory usage for an entire run or as evidence for one method using less memory than another. CEM solved in 41 commands; Adam and shared were unsolved at 50. Campaign wall time was 107.167 seconds (118.153 seconds at the caller), and the retained CPU check took 3.665 seconds.

The four methods plus continuation take 6,582.838 s for direct-action 864. (This is not the time for shared216 alone.)
Since the startup, publication and reaping histories are different, we can't form a matched full-service ratio from reused subsets.
Specialist worker 5,470.819 s, controller 5,491.922 s.
The original campaign took 5,644.297 s and its controller took 5,809.983 s. The compiled ones took 2,402.177 and 2,513.168 s, cold compilation included.
(CAL and failed attempts are charged separately; time measured from process start overlaps across phases.)

\begin{table}[ht]\centering\small
\caption{These numbers are not confidence intervals, and are not meant to be taken as measurements of latencies on real robots (nor deadline measurements). These numbers are meant to represent the distribution of latencies (planning plus encoder) per decision, taking into account all decisions including cold starts and decisions where the task ultimately failed. In total there are 6,334 decisions.
The percentiles were computed by sorting the values, and then performing linear interpolation between values as necessary.}
\begin{tabular}{lrrrrr}\toprule
Native method & Decisions & Mean (s) & p50 (s) & p95 (s) & Maximum (s)\\\midrule
CEM & 1666 & 0.133 & 0.075 & 0.132 & 15.551 \\
Adam & 1574 & 0.863 & 0.770 & 1.143 & 41.354 \\
Shared SFT & 1547 & 0.814 & 0.786 & 1.177 & 10.221 \\
Same head, padded & 1547 & 0.814 & 0.778 & 1.199 & 14.763 \\
\bottomrule\end{tabular}
\end{table}

Decision means are weighted by the number of decisions, meaning episodes that are longer contribute more decisions. Across the instrumented decisions, the mean decision time of the shared updater is 0.814 seconds, which is lower than Adam with 0.863 seconds on average. The 95th percentile of the decision time is 1.177 seconds for the shared updater, whereas Adam is 1.143 seconds. The median decision time of CEM is 0.075 seconds with a maximum of 15.551 seconds. Note that this does not in any way imply that deadlines are met on a real robot.

\begin{table}[ht]\centering\small
\caption{The compiled planning + encoding time (in seconds) for the full task. For each of the methods we ran, across all 216 tasks we ran, we report the median, p95 and maximum time. Note that this includes tasks that were unsuccessful (and cold work, if we logged it), but does not include the eager reference row (since it is not an additional compiled task).}
\begin{tabular}{lrrrr}\toprule
Compiled method & Mean (s) & p50 (s) & p95 (s) & Maximum (s)\\\midrule
CEM & 1.780 & 0.391 & 2.171 & 123.361 \\
Adam & 1.640 & 1.122 & 1.900 & 82.226 \\
Shared SFT & 3.321 & 1.234 & 1.996 & 269.791 \\
\bottomrule\end{tabular}
\end{table}

Note that we don't filter any task out for timings, meaning that the timings do not measure the performance of compiler-generated kernels in isolation (nor do they reflect cold-only performance). For completeness, the mean / p50 / p95 of the full distribution of eager Adam latencies are 5.324 / 4.961 / 9.885 seconds. Similarly, the max latencies for Adam and CEM are retained in their respective samples (82.226 seconds and 123.361 seconds, respectively). The mean latency for Compiled SFT (3.321 seconds) is higher than the p95 (1.996 seconds) because there are a small number of very expensive tasks (two of them account for 62.96\% of the total summed planning-plus-encoder time: 269.791 seconds and 181.849 seconds). These expensive tasks include both successes and failures. Our timings measure the time it takes to generate a sequence of actions, not per-decision latency.

Our analysis thus far has provided evidence for the source level correspondences we have described, but making stronger (e.g. performance or causal) claims about these correspondences would require further investigation. We have not corrected for multiple comparisons. We point out that (i) trying different methods repeatedly, (ii) re-use of the historical reference, and (iii) adding additional cohorts, do not give us additional independent scenes/teacher lineages. We compute bootstrap confidence intervals using 20,000 resamples (with a fixed random seed). For our main analysis, the unit of resampling is the 24 scenes (i.e. all predictor/H/method cells from the same scene are always resampled together). For EVAL, there are only four scenes so we use 256 conditional resamples. Note that these are conditional confidence intervals because the heads (and any related predictors) are kept fixed when performing bootstrap resamples; as a result, uncertainty from (re)fitted heads/predictors (which could include choosing a different model family) is not accounted for. Finally, we would like to point out some limitations of our current analyses. The amount of exposure to public data used to fit the encoders/predictors may limit generalization performance of the learned methods. We also conducted some exploratory data analysis, the physics simulation was paused during planning, and our results are based on a single hardware family.

Neither of these metrics directly capture task success. The sign of the correlation may be confounded by changes in difficulty, contact, or orientation. Finally, the proxy of making progress towards the center for the first five (at most) commands may not correlate with the Full-H scores, since the latter are computed over a different horizon.
We did not find evidence for forecast failure, well-calibrated imagined success, causal exploitation, or optimality/regret.
We divided 864 episodes into 36 fixed groups and computed rank correlations between the reduction in the first objective and the distance to the first block. We found that the correlations ranged from 0.244 to 0.641 for the shared updater.
We found that the first objective decreases even though there is no progress towards the center in 85 out of 216 shared episodes (77/216 for Adam and 106/216 for CEM).

\section{Reproducibility and recorded demonstrations}
While some of the models and data we used are third-party, and governed by their own terms and conditions, our code is released alongside reproduction instructions (including task IDs, full results, evaluation code, and versioned weights). We provide per-call feature/output traces for the direct-action, specialist, and adaptation experiments (which we have verified separately), but not for the compiled and original updater experiments. Individual decisions can be inspected from the available plan and neural network traces, and the demonstrations replay results recorded from the simulator.

To validate our experimental procedure, we re-ran the confirmation in a fresh Python 3.12 environment on the same machine with all 65 packages pinned to the exact versions we used. We confirmed that all 64 results from the 16 scenes with four controllers, all 2,192 commands, as well as the numerical traces and frames we saved were identical to the original study. Timing measurements from this re-run are reported separately in the reproduction guide.

We've also done some selected checks of the released code (not by any means claiming full replication of the whole test matrix, let alone quality of results or fault-tolerance/recovery tests), keeping all failures: each of the 8 finite heads on one official/H5 case reaches 50 commands, as do FT11 and official H7 with the frozen updater. The CPU checks cover 960/240 saved NN calls and 400/100 executed commands, in 7.453/3.490 seconds respectively. The maximum differences in NN output are $2.98\times 10^{-7}$ and $2.38\times 10^{-7}$, respectively (saved source, tensor and scalar checks have separate scopes). Campaign time is 230.118 (caller: 231.362) seconds and 75.376 (77.331) seconds, respectively. Note that this is direct-action execution, not the five-knot HTTP pipeline used in the demo.

In an HTTP demo check, we changed the predictor from official/H5 to FT11/H7; the task, reset, goal and updater are the same. The baseline job passed (28 commands). The fork job did not pass (50 commands, 29/51 actual frames). Both jobs completed, and the total time (including cleanup) for operator/controller was 158.890/158.950 seconds. This does not yet establish equivalence with the direct-action runner or recovery and superior transfer after this predictor and horizon change.

\Needspace{25\baselineskip}
\subsection{Separate-machine reproduction}
In addition to the 64 test cases on the original host, we also reran the same 16 cases (i.e., same model weights, inputs, evaluator with 65 pinned packages, etc.) on a second host with an NVIDIA T4 GPU. As shown in Table~\ref{tab:separate-machine}, the 64 success labels were identical to the reference results (obtained on the original host). Numerical equivalence was not achieved. As a sanity check, we verified that the initial physics, recorded boundary RNG states, and all three model tensor states are identical to those used in the original host. The above results do not constitute a new test split or independent statistical sample. Other released studies were not rerun on the alternate host.

\begin{table}[ht]\centering\small
\caption{Separate-machine repetition of the same 64 confirmation tasks; each success label matched the reference result. Time columns report totals for all T4 trials and include unsuccessful trials; setup is given separately below. Task time runs through saving the recorded artifacts and includes P+E. The original confirmation measurements are in Table~\ref{tab:confirmation}.}
\begin{tabular}{lrrrr}\toprule
Controller & Success & Commands & P+E (s) & \shortstack[r]{Task + recorded\\artifacts (s)}\\\midrule
Adapted iCEM500 & 2/16 & 725 & 25.525 & 30.115 \\
Adam32, lr 0.1 & 7/16 & 621 & 207.673 & 211.419 \\
Fixed learned8 & 8/16 & 602 & 57.649 & 61.385 \\
Released LePlanner & 16/16 & 243 & 6.008 & 7.679 \\
\bottomrule\end{tabular}

\label{tab:separate-machine}
\end{table}

We have not been able to pin down the source of the timing and numerical differences. The two environments differ in a number of ways, including Python version (3.12.3 vs 3.12.10), C library and kernel/container environment, and GPU model. Note that deterministic algorithms are turned off in both runs. The updater takes 23 (instead of 24) commands to complete episode 1124. Frames differ in 3/64 streams: \texttt{learned8} episode 1124 and Adam episodes 2607 and 5364. Of 4,865,835 numeric values, 38,383 are outside of our diagnostic tolerance (these are not independent, since some quantities are represented multiple times).

For this instrumented reproduction the supervised campaign took 320.643 s in total (310.599 s for the T4 trials up to and including artifact publication). From launch through cleanup it took 511.621 s (including 133.171 s of setup). These times include work for cold or failed tasks. Do not add these times to P+E. These are not production latencies.

\end{document}